\pdfoutput=1

\documentclass[11pt]{article}

\usepackage[]{acl}

\usepackage{times}
\usepackage{latexsym}

\usepackage{graphicx}
\usepackage{amsmath}
\usepackage{booktabs}
\usepackage{amsfonts,amssymb}
\usepackage{amssymb}
\usepackage{url}
\usepackage{enumitem}
\usepackage{multirow}
\usepackage{amsthm}
\usepackage{tikz}
\usepackage{tikz-dependency}
\usepackage{subcaption}

\definecolor{graycolor}{RGB}{120,120,120}
\newcommand{\gray}[1]{\textcolor{graycolor}{#1}}

\usepackage[T1]{fontenc}

\usepackage[utf8]{inputenc}

\usepackage{microtype}

\title{FormNetV2: Multimodal Graph Contrastive Learning for \\Form Document Information Extraction}

\author{
\normalsize
Chen-Yu Lee$^{1}$\thanks{\ All work done at Google. Correspondence to: Chen-Yu Lee <chenyulee@google.com>, Chun-Liang Li <chunliang@google.com>}, Chun-Liang Li$^{1}$, Hao Zhang$^2$, Timothy Dozat$^2$, Vincent Perot$^2$, \\
\bf
\normalsize
Guolong Su$^2$, Xiang Zhang$^{1}$, Kihyuk Sohn$^2$, Nikolai Glushnev$^{3}$, Renshen Wang$^2$, \\
\bf
\normalsize
Joshua Ainslie$^2$, Shangbang Long$^2$, Siyang Qin$^2$, Yasuhisa Fujii$^2$, Nan Hua$^2$, Tomas Pfister$^{1}$ \\
\normalsize
$^1$Google Cloud AI Research,
$^2$Google Research,
$^3$Google Cloud AI
}

\begin{document}
\maketitle
\begin{abstract}
The recent advent of self-supervised pre-training techniques has led to a surge in the use of multimodal learning in form document understanding. However, existing approaches that extend the mask language modeling to other modalities require careful multi-task tuning, complex reconstruction target designs, or additional pre-training data. In FormNetV2, we introduce a centralized multimodal graph contrastive learning strategy to unify self-supervised pre-training for all modalities in one loss. The graph contrastive objective maximizes the agreement of multimodal representations, providing a natural interplay for all modalities without special customization. In addition, we extract image features within the bounding box that joins a pair of tokens connected by a graph edge, capturing more targeted visual cues without loading a sophisticated and separately pre-trained image embedder. FormNetV2 establishes new state-of-the-art performance on FUNSD, CORD, SROIE and Payment benchmarks with a more compact model size.
\end{abstract}
\vspace{5mm}

\section{Introduction}
Automated information extraction is essential for many practical applications, with form-like documents posing unique challenges compared to article-like documents, which have led to an abundance of recent research in the area. 
In particular, form-like documents often have complex layouts that contain structured objects like tables, columns, and fillable regions. 
Layout-aware language modeling has been critical for many successes~\cite{xu2020layoutlm,majumder2020representation,lee2022formnet}.

To further boost the performance, many recent approaches adopt multiple modalities~\cite{xu2020layoutlmv2,huang2022layoutlmv3,appalaraju2021docformer}. 
Specifically, the image modality adds more structural information and visual cues to the existing layout and text modalities.
They therefore extend the masked language modeling (MLM) from text to masked image modeling (MIM) for image and text-image alignment (TIA) for cross-modal learning.
The alignment objective may also help to prime the layout modality, though it does not directly involve text layouts or document structures.

In this work, we propose FormNetV2, a multimodal transformer model for form information extraction.
Unlike existing works -- which may use the whole image as one representation~\cite{appalaraju2021docformer}, or image patches~\cite{xu2020layoutlmv2}, or image features of token bounding boxes~\cite{xu2020layoutlm} -- we propose using image features extracted from the region bounded by a \emph{pair} of tokens connected in the constructed graph. 
This allows us to capture a richer and more targeted visual component of the intra- and inter-entity information. 
Furthermore, instead of using multiple self-supervised objectives for each individual modality, we introduce graph contrastive learning~\cite{li2019graph,you2020graph,zhu2021empirical} to learn multimodal embeddings jointly.    
These two additions to FormNetV1~\cite{lee2022formnet} enable the graph convolutions to produce better super-tokens, resulting in both improved performance and a smaller model size.

In experiments, FormNetV2 outperforms its predecessor FormNetV1 as well as the existing multimodal approaches on four standard benchmarks.
In particular, compared with FormNetV1, FormNetV2 outperforms it by a large margin on FUNSD (86.35 v.s. 84.69) and Payment (94.90 v.s. 92.19); compared with DocFormer~\cite{appalaraju2021docformer}, FormNetV2 outperforms it on FUNSD and CORD with nearly 2.5x less number of parameters.

\section{Related Work}
Early works on form document information extraction are based on rule-based models or learning-based models with handcrafted features~\cite{lebourgeois1992fast,o1993document,ha1995recursive,simon1997fast,marinai2005artificial,chiticariu2013rule}. Later on, various deep neural models have been proposed, including methods based on recurrent nets~\cite{palm2017cloudscan,aggarwal2020form2seq}, convolutional nets~\cite{katti2018chargrid,zhao2019cutie,denk2019bertgrid}, and transformers~\cite{majumder2020representation,garncarek2020lambert,wang2022queryform}.

Recently, in addition to the text, researchers have explored the layout attribute in form document modeling, such as the OCR word reading order~\cite{lee2021rope,gu2022xylayoutlm}, text coordinates~\cite{majumder2020representation,xu2020layoutlm,garncarek2020lambert,li2021structurallm,lee2022formnet}, layout grids~\cite{lin2021vibertgrid}, and layout graphs~\cite{lee2022formnet}. 
The image attribute also provides essential visual cues such as fonts, colors, and sizes. Other visual signals can be useful as well, including logos and separating lines from form tables. \citet{xu2020layoutlm} uses Faster R-CNN~\cite{ren2015faster} to extract token image features; \citet{appalaraju2021docformer} uses ResNet50~\cite{he2016deep} to extract full document image features; \citet{li2022dit} use ViT~\cite{dosovitskiy2020image} with FPN~\cite{lin2017feature} to extract non-overlapping patch image features. These sophisticated image embedders require a separate pre-training step using external image datasets (e.g. ImageNet~\cite{russakovsky2015imagenet} or PubLayNet~\cite{zhong2019publaynet}), and sometimes depend upon a visual codebook pre-trained by a discrete variational auto-encoder (dVAE).

When multiple modalities come into play, different supervised or self-supervised multimodal pre-training techniques have been proposed.
They include mask prediction, reconstruction, and matching for one or more modalities~\cite{xu2020layoutlm,xu2020layoutlmv2,appalaraju2021docformer,li2021selfdoc,gu2022unified,huang2022layoutlmv3,li2022dit,pramanik2020towards}. Next-word prediction~\cite{kim2022ocr} or length prediction~\cite{li2021structext} have been studied to bridge text and image modalities. Direct and relative position predictions~\cite{cosma2020self,wei2020robust,li2021structurallm,wang2022lilt,li2021structext} have been proposed to explore the underlying layout semantics of documents. Nevertheless, these pre-training objectives require strong domain expertise, specialized designs, and multi-task tuning between involved modalities. In this work, our proposed graph contrastive learning performs multimodal pre-training in a centralized design, unifying the interplay between all involved modalities without the need for prior domain knowledge.

\begin{figure*}[t!]
    \centering
    \includegraphics[width=0.9\linewidth]{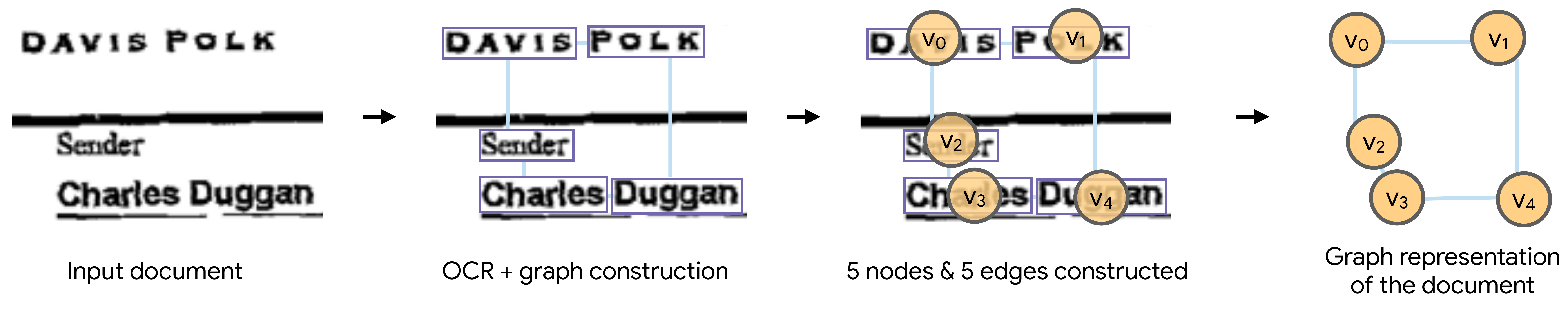}
    \vspace{-4mm}
    \caption{Graph of a sample region from a form. Token bounding boxes are identified, and from them the graph is constructed. Nodes are labeled and the graph structure is shown abstracted away from its content.}
    \label{fig:graph_construction}
    \vspace{-3mm}
\end{figure*}

\begin{figure}[t!]
    \centering
    \includegraphics[width=0.9\linewidth]{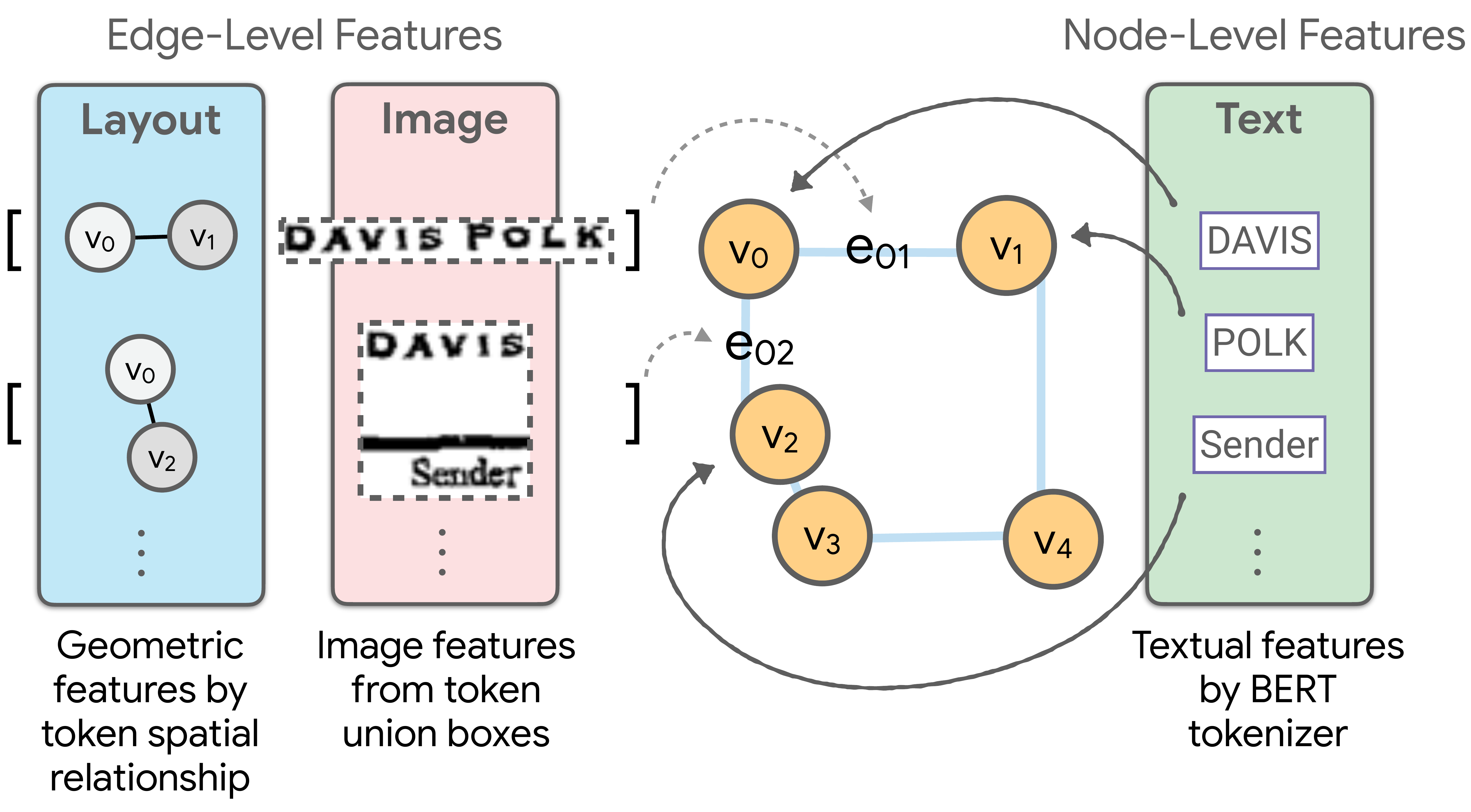}
    \vspace{-4mm}
    \caption{Multimodal graph representations are composed from three modalities: text at node-level; concatenation of layout and image at edge-level.}
    \label{fig:multimodal_feat}
    \vspace{-3mm}
\end{figure}

\section{FormNetV2}
We briefly review the backbone architecture FormNetV1~\cite{lee2022formnet} in Sec~\ref{sec:preliminaries}, introduce the multimodal input design in Sec~\ref{sec:multimodal_input}, and detail the multimodal graph contrastive learning in Sec~\ref{alg:gcl}.

\subsection{Preliminaries}
\label{sec:preliminaries}
\paragraph{ETC.}
FormNetV1~\cite{lee2022formnet} uses Extended Transformer Construction~\citep[ETC;][]{ainslie2020etc} as the backbone to work around the quadratic memory cost of attention for long form documents. ETC permits only a few special tokens to attend to every token in the sequence (global attention); all other tokens may only attend to $k$ local neighbors within a small window, in addition to these special tokens (local attention). This reduces the computational complexity from $O(n^2)$ query-key pairs that need scoring to $O(kn)$. Eq.\ (\ref{local attention}) formalizes the computation of the attention vector $\mathbf{a}_0$ for a model with one global token at index 0, and Eq.\ (\ref{local attention}) formalizes computation of the attention vector $\mathbf{a}_{i > 0}$ for the rest of the tokens in the model.
\begin{align}
    \label{global attention}\mathbf{a}_0 &= \texttt{attend}(\mathbf{h}_0, [\mathbf{h}_0, \mathbf{h}_1, \ldots, \mathbf{h}_n])\\
    \label{local attention}\mathbf{a}_{i > 0} &= \texttt{attend}(\mathbf{h}_i, [\mathbf{h}_0, \mathbf{h}_{i - k}, \ldots, \mathbf{h}_{i + k}])
\end{align}

\paragraph{Rich Attention.}
To address the distorted semantic relatedness of tokens created by imperfect OCR serialization, FormNetV1 adapts the attention mechanism to model spatial relationships between tokens by proposing Rich Attention, a mathematically sound way of conditioning attention on low-level spatial features without resorting to quantizing the document into regions associated with distinct embeddings in a lookup table. In Rich Attention, the model constructs the (pre-softmax) attention score (Eq.\ \ref{rich attention score}) from multiple components: the usual transformer attention score (Eq.\ \ref{transformer score}); the order of tokens along the x-axis and the y-axis (Eq.\ \ref{order score}); and the log distance (in number of pixels) between tokens, again along both axes (Eq.\ \ref{distance score}). The expression for a transformer head with Rich Attention on the x-axis is provided in Eqs.\ (\ref{order}--\ref{rich attention score}); we refer the interested reader to \citet{lee2022formnet} for further details.
\begin{align}
    \label{order}o_{ij} &= \texttt{int}(x_i < x_j)\\
    \label{distance}d_{ij} &= \ln(1 + |x_i - x_j|)\\
    \label{ideal order}p_{ij} &= \texttt{Sigmoid}(\text{affine}^{(p)}([\mathbf{q}_i; \mathbf{k}_j]))\\
    \label{ideal distance}\mu_{ij} &= \text{affine}^{(\mu)}([\mathbf{q}_i; \mathbf{k}_j])\\
    \label{transformer score}s^{(t)}_{ij} &= \mathbf{q}_i^\top\mathbf{k}_j\\
    \label{order score}s^{(o)}_{ij} &= o_{ij}\ln(p_{ij}) + (1-o_{ij})\ln(1-p_{ij})\\
    \label{distance score}s^{(d)}_{ij} &= -\frac{\theta^2(d_{ij} - \mu_{ij})^2}{2}\\
    \label{rich attention score}s_{ij} &= s_{ij}^{(t)} + s_{ij}^{(o)} + s_{ij}^{(d)}
\end{align}

\paragraph{GCN.} Finally, FormNetV1 includes a graph convolutional network (GCN) contextualization step \emph{before} serializing the text to send to the ETC transformer component. The graph for the GCN locates up to $K$ neighbors for each token -- defined broadly by geographic ``nearness'' -- before convolving their token embeddings to build up supertoken representations as shown in Figure~\ref{fig:graph_construction}. This allows the network to build a weaker but more complete picture of the layout modality than Rich Attention, which is constrained by local attention.

\begin{figure}
\centering
\begin{subfigure}{.2\textwidth}
  \centering
  \includegraphics[width=0.9\linewidth]{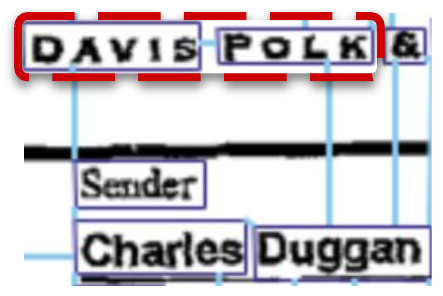}
  \vspace{-1mm}
  \caption{Within entity}
  \label{fig:img_sub1}
\end{subfigure}%
\hspace{6mm}
\begin{subfigure}{.2\textwidth}
  \centering
  \includegraphics[width=0.9\linewidth]{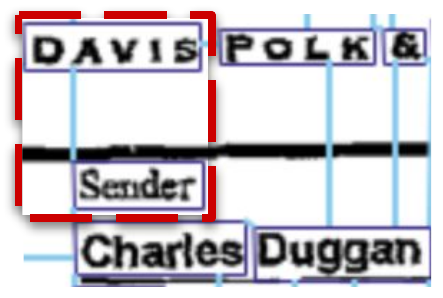}
  \vspace{-1mm}
  \caption{Cross entity}
  \label{fig:img_sub2}
\end{subfigure}
\vspace{-2mm}
\caption{Image features are extracted from bounding boxes (red) that join pairs of tokens connected by edges to capture (a) similar patterns within an entity, or (b) dissimilar patterns or separating lines between entities.}
\label{fig:image_modality}
\vspace{-3mm}
\end{figure}

The final system was pretrained end-to-end with a standard masked language modeling (MLM) objective. See Sec~\ref{app:preliminaries} in Appendix for more details.

\subsection{Multimodal Input}
\label{sec:multimodal_input}
In FormNetV2, we propose adding the image modality to the model in addition to the text and layout modalities that are already used in FormNetV1 (Sec 3.3 in~\citet{lee2022formnet}). We expect that image features from documents contain information absent from the text or the layout, such as fonts, colors, and sizes of OCR words. 

To do this, we run a ConvNet to extract dense image features on the whole document image, and then use Region-of-Interest (RoI) pooling~\cite{he2017mask} to pool the features within the bounding box that joins a pair of tokens connected by a GCN edge. Finally, the RoI pooled features go through another small ConvNet for refinement. After the image features are extracted, they are injected into the network through concatenation with the existing layout features at edges of the GCN. Figure~\ref{fig:multimodal_feat} illustrates how all three modalities are utilized in this work and Sec~\ref{sec:setup} details the architecture.

\begin{figure*}[t!]
    \centering
    \includegraphics[width=0.95\linewidth]{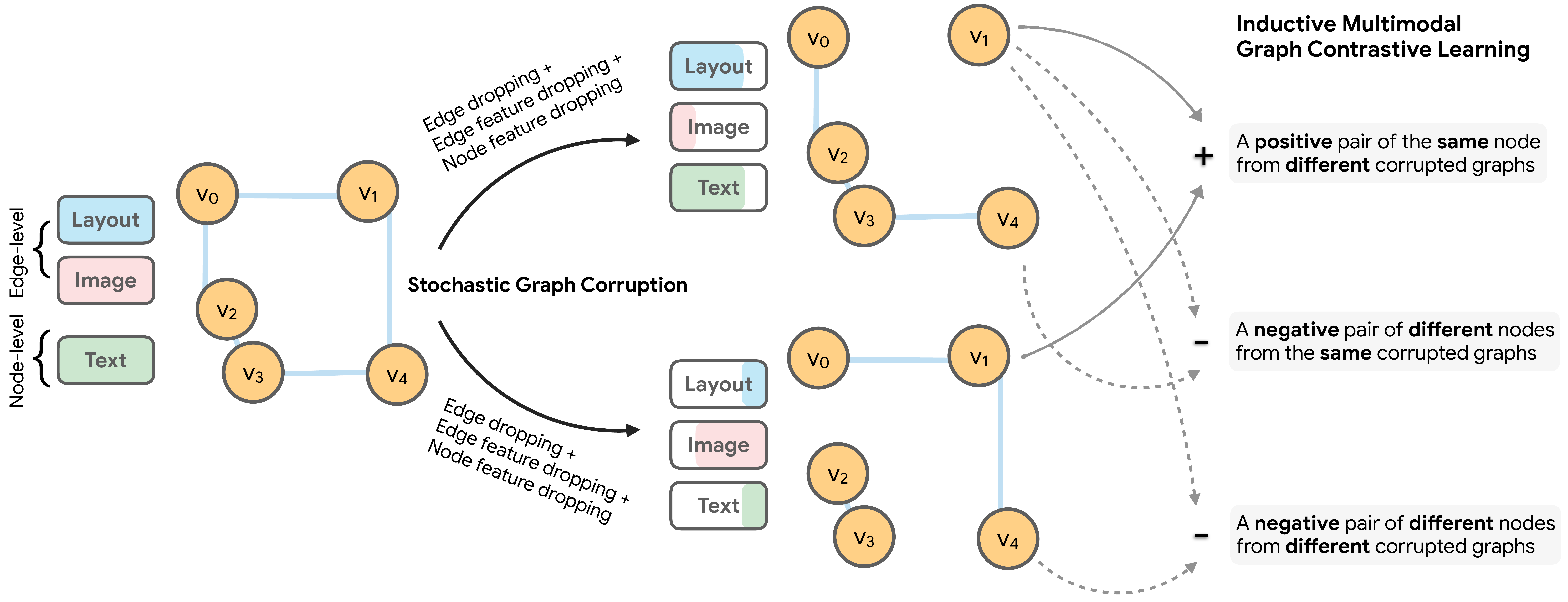}
    \vspace{-4mm}
    \caption{Multimodal graph contrastive learning. Two corrupted graphs are sampled from an input graph by corruption of graph topology (edges) and attributes (multimodal features). The system is trained to identify which pair of nodes across all pairs of corrupted nodes (including within the same graph) came from the same node.}
    \label{fig:multimodal_gcl}
    \vspace{-2mm}
\end{figure*}

Most of the recent approaches (Table~\ref{overall_comparison}) that incorporate image modality extract features from either (a) the whole image as one vector, (b) non-overlapping image patches as extra input tokens to transformers, or (c) token bounding boxes that are added to the text features for all tokens.

However, form document images often contain OCR words that are relatively small individually and are densely distributed in text blocks. They also contain a large portion of the background region without any texts. Therefore, the aforementioned method (a) only generates global visual representations with large noisy background regions but not targeted entity representations; method (b) tends to be sensitive to the patch size and often chops OCR words or long entities to different patches, while also increasing computational cost due to the increased token length; and method (c) only sees regions within each token's bounding box and lacks context between or outside of tokens.

On the other hand, the proposed edge-level image feature representation can precisely model the relationship between two nearby, potentially related ``neighbor'' tokens and the surrounding region, while ignoring all irrelevant or distracting regions. Figure~\ref{fig:image_modality} demonstrates that the targeted RoI image feature pooling through the union bounding box can capture any similar patterns (e.g. font, color, size) within an entity (left) or dissimilar patterns or separating lines between entities (right). See Sec~\ref{sec:ablation} for detailed discussion.

\subsection{Multimodal Graph Contrastive Learning}
\label{alg:gcl}

Previous work in multimodal document understanding requires manipulating multiple supervised or self-supervised objectives to learn embeddings from one or multiple modalities during pre-training. By contrast, in FormNetV2, we propose utilizing the graph representation of a document to learn multimodal embeddings with a contrastive loss.

Specifically, we first perform stochastic graph corruption to sample two corrupted graphs from the original input graph of each training instance. This step generates node embeddings based on partial contexts. Then, we apply a contrastive objective by maximizing agreement between tokens at node-level. That is, the model is asked to identify which pairs of nodes across all pairs of nodes -- within the same graph and across graphs -- came from the same original node. We adopt the standard normalized temperature-scaled cross entropy (NT-Xent) loss formulation~\cite{chen2020simple, wu2018unsupervised, oord2018representation, sohn2016improved} with temperature 0.1 in all experiments.

To build a centralized contrastive loss that unifies the interactions between multiple input modalities, we corrupt the original graph at both graph topology level and graph feature level. Topology corruption includes edge dropping by randomly removing edges in the original graph. Feature corruption includes applying dropping to all three modalities: dropping layout and image features from edges and dropping text features from nodes. Note that we only corrupt the graph in the GCN encoder and keep the ETC decoder intact to leverage the semantically meaningful graph representation of the document during graph contrastive learning.

To further diversify the contexts in two corrupted graphs and reduce the risk of training the model to over-rely on certain modalities, we further design an inductive graph feature dropping mechanism by adopting imbalanced drop-rates of modalities between the two corrupted graphs. Precisely, for a given modality, we discard $p$ percent of the features in the first corrupted graph and discard $1-p$ percent of the features in the second corrupted graph. Experiments in Sec~\ref{sec:ablation} show that $p=0.8$ works best empirically and the inductive feature dropping mechanism provides further performance boost over the vanilla version. We stipulate that this boom-and-bust approach to regularization allows the model to learn rich, complex representations that take full advantage of the model's capacity without becoming overly dependent on specific feature interactions. Figure~\ref{fig:multimodal_gcl} illustrates the overall process.

The proposed graph contrastive objective is also general enough in principle to adopt other corruption mechanisms~\cite{zhu2020deep, hassani2020contrastive, you2020graph, velickovic2019deep}. The multimodal feature dropping provides a natural playground to consume and allow interactions between multiple input modalities in one single loss design. It is straightforward to extend the framework to include more modalities without the need for hand crafting specialized loss by domain experts. To the best of our knowledge, we are the first to use graph contrastive learning during pre-training for form document understanding.

\section{Evaluation}
\subsection{Datasets}

\paragraph{FUNSD.}
FUNSD~\cite{jaume2019} contains a collection of research, marketing, and advertising forms that vary extensively in their structure and appearance.
The dataset consists of 199 annotated forms with 9,707 entities and 31,485 word-level annotations for 4 entity types: header, question, answer, and other.
We use the official 75-25 split for the training and test sets.

\vspace{-2mm}
\paragraph{CORD.}
CORD~\cite{park2019cord} contains over 11,000 Indonesian receipts from shops and restaurants.
The annotations are provided in 30 fine-grained semantic entities such as store name, quantity of menu, tax amount, discounted price, etc. 
We use the official 800-100-100 split for training, validation, and test sets.

\vspace{-2mm}
\paragraph{SROIE.}
The ICDAR 2019 Challenge on Scanned Receipts OCR and key Information Extraction (SROIE)~\cite{huang2019icdar2019} 
offers 1,000 whole scanned receipt images and annotations. 626 samples are for training and 347 samples are for testing. The task is to extract four predefined entities: company, date, address, or total.

\vspace{-2mm}
\paragraph{Payment.}
We use the large-scale payment data \cite{majumder2020representation} that consists of roughly 10,000 documents and 7 semantic entity labels from human annotators.
We follow the same evaluation protocol and dataset splits used in~\citet{majumder2020representation}.

\subsection{Experimental Setup}
\label{sec:setup}

We follow the FormNetV1~\cite{lee2022formnet} architecture with a slight modification to incorporate multiple modalities used in the proposed method. Our backbone model consists of a 6-layer GCN encoder to generate structure-aware super-tokens, followed by a 12-layer ETC transformer decoder equipped with Rich Attention for document entity extraction. The number of hidden units is set to 768 for both GCN and ETC. The number of attention heads is set to 1 in GCN and 12 in ETC. The maximum sequence length is set to 1024. We follow~\citet{ainslie2020etc, lee2022formnet} for other hyper-parameter settings. For the image embedder architecture, see Sec~\ref{app:image_embedder} in Appendix.

\begin{table*}[!ht]
\setlength{\tabcolsep}{6pt} 
\centering
\resizebox{1\textwidth}{!}{
\footnotesize{
\begin{tabular}{llccccccc}
\toprule
\textbf{Dataset} & \textbf{Method} & \textbf{P} & \textbf{R} & \textbf{F1} & \gray{\textbf{F1}\textsuperscript{\textdagger}} & \textbf{Modality} & \textbf{Image Embedder} & \textbf{\#Params}    \\
\toprule
FUNSD & SPADE~\cite{hwang2020spatial} & - & - & 70.5 & - & T+L & - & 110M  \\
& UniLMv2~\cite{bao2020unilmv2} & 67.80 & 73.91 & 70.72 & - & T & - & 355M \\
& LayoutLMv1~\cite{xu2020layoutlm} & 75.36 & 80.61 & 77.89 & - & T+L & - & 343M  \\
& DocFormer~\cite{appalaraju2021docformer} & 81.33 & 85.44 & 83.33 & - & T+L+I & ResNet50 & 502M \\
& FormNetV1~\cite{lee2022formnet} & 85.21 & 84.18 & 84.69 & - & T+L & - & 217M \\
\cmidrule{2-9}
& LayoutLMv1~\cite{xu2020layoutlm} & 76.77 & 81.95 & 79.27 & - & T+L+I & ResNet101 & 160M  \\
& LayoutLMv2~\cite{xu2020layoutlmv2} & 83.24 & 85.19 & 84.20 & - & T+L+I &  ResNeXt101-FPN & 426M  \\
& DocFormer~\cite{appalaraju2021docformer} & 82.29 & 86.94 & 84.55 & - & T+L+I & ResNet50 & 536M  \\
& StructuralLM~\cite{li2021structurallm} & - & - & - & \gray{85.14} & T+L & - & 355M  \\
& LayoutLMv3~\cite{huang2022layoutlmv3} & 81.35 & 83.75 & 82.53 & \gray{92.08} & T+L+I & Tokenization & 368M  \\
\cmidrule{2-9}
& FormNetV2 (ours) & 85.78 & 86.94 & \textbf{86.35} & \gray{\textbf{92.51}} & T+L+I & 3-layer ConvNet & 204M \\
\midrule
CORD & SPADE~\cite{hwang2020spatial} & - & - & 91.5 & - & T+L & - & 110M  \\
& UniLMv2~\cite{bao2020unilmv2} & 91.23 & 92.89 & 92.05 & - & T  & - & 355M \\
& LayoutLMv1~\cite{xu2020layoutlmv2} & 94.32 & 95.54 & 94.93 & - & T+L & -  & 343M \\
& DocFormer~\cite{appalaraju2021docformer} & 96.46 & 96.14 & 96.30 & - & T+L+I & ResNet50 & 502M \\
& FormNetV1~\cite{lee2022formnet} & 98.02 & 96.55 & 97.28 & - & T+L & - & 345M \\
\cmidrule{2-9}
& LayoutLMv2~\cite{xu2020layoutlmv2} & 95.65 & 96.37 & 96.01 & - & T+L+I &  ResNeXt101-FPN & 426M \\
& TILT~\cite{powalski2021going} & - & - & 96.33 & - & T+L+I & U-Net & 780M \\
& DocFormer~\cite{appalaraju2021docformer} & 97.25 & 96.74 & 96.99 & - & T+L+I & ResNet50 & 536M \\
& LayoutLMv3~\cite{huang2022layoutlmv3} & 95.82 & 96.03 & 95.92 & \gray{97.46} & T+L+I & Tokenization & 368M  \\
\cmidrule{2-9}
& FormNetV2 (ours) & 97.74 & 97.00 & \textbf{97.37} & \gray{\textbf{97.70}} & T+L+I & 3-layer ConvNet & 204M \\
\midrule
SROIE
& UniLMv2~\cite{bao2020unilmv2} & - & - & 94.88 & - & T  & - & 355M \\
& LayoutLMv1~\cite{xu2020layoutlmv2} & 95.24 & 95.24 & 95.24 & - & T+L & -  & 343M \\
& LayoutLMv2~\cite{xu2020layoutlmv2} & 99.04 & 96.61 & 97.81 & - & T+L+I &  ResNeXt101-FPN & 426M \\
\cmidrule{2-9}
& FormNetV2 (ours) & 98.56 & 98.05 & \textbf{98.31} & - & T+L+I & 3-layer ConvNet & 204M \\
\midrule
Payment & NeuralScoring~\cite{majumder2020representation} & - & - & 87.80 & - & T+L & - & - \\
& FormNetV1~\cite{lee2022formnet}   & 92.70 & 91.69 & 92.19 & - & T+L & - & 217M \\
\cmidrule{2-9}
& FormNetV2 (ours) & 94.11 & 95.71 & \textbf{94.90} & - & T+L+I & 3-layer ConvNet & 204M \\
\bottomrule
\end{tabular}
}
}
\vspace{-3mm}
\caption{\label{overall_comparison} Entity-level precision, recall, and F1 score comparisons on four standard benchmarks. ``T/L/I'' denotes ``text/layout/image'' modality. The proposed FormNetV2 establishes new state-of-the-art results on all four datasets.
FormNetV2 significantly outperforms the most recent DocFormer~\cite{appalaraju2021docformer} and LayoutLMv3~\cite{huang2022layoutlmv3} while using a 38\% and 55\% sized model, respectively. Note that LayoutLMv3~\cite{huang2022layoutlmv3} and StructuralLM~\cite{li2021structurallm} use segment-level layout positions that incorporate ground truth entity bounding boxes, which is less practical for real-world applications. We nevertheless report our results under the same protocol in column \gray{\textbf{F1}\textsuperscript{\textdagger}}. See Sec~\ref{sec:benchmark} and Sec~\ref{app:v3_ablation} in Appendix for details.}
\vspace{-5mm}
\end{table*}

\vspace{-2mm}
\paragraph{Pre-training.}
We pre-train FormNetV2 using two unsupervised objectives: Masked Language Modeling (MLM)~\cite{taylor1953cloze, devlin2018bert} and the proposed multimodal Graph Contrastive Learning (GCL). 

Different from BERT~\cite{devlin2018bert}, here MLM has access to layout and image modalities during pre-training similar to~\citet{appalaraju2021docformer, xu2020layoutlmv2, xu2020layoutlm}. Nevertheless, the layout and image features are constructed at edge level instead of at node level, supplementing the text features for better underlying representation learning without directly leaking the trivial information.

GCL provides a natural playground for effective interactions between all three modalities from a document in a contrastive fashion. For each graph representation of a document, we generate two corrupted views by edge dropping, edge feature dropping, and node feature dropping with dropping rates \{0.3, 0.8, 0.8\}, respectively. The weight matrices in both GCN and ETC are shared across the two views.

We follow~\citet{appalaraju2021docformer, xu2020layoutlmv2, xu2020layoutlm} and use the large-scale IIT-CDIP document collection~\cite{lewis2006building} for pre-training, which contains 11 million document images. We train the models from scratch using Adam optimizer with batch size of 512. The learning rate is set to 0.0002 with a warm-up proportion of 0.01. We find that GCL generally converges faster than MLM, therefore we set the loss weightings to 1 and 0.5 for MLM and GCL, respectively.

Note that we do not separately pre-train or load a pre-trained checkpoint for the image embedder as done in other recent approaches shown in Table~\ref{overall_comparison}. In fact, in our implementation, we find that using sophisticated image embedders or pre-training with natural images, such as ImageNet~\cite{russakovsky2015imagenet}, do not improve the final downstream entity extraction F1 scores, and they sometimes even degrade the performance. This might be because the visual patterns presented in form documents are drastically different from natural images that have multiple real objects. The best practice for conventional vision tasks (classification, detection, segmentation) might not be optimal for form document understanding.

\vspace{-2mm}
\paragraph{Fine-tuning.}
We fine-tune all models for the downstream entity extraction tasks in the experiments using Adam optimizer with batch size of 8. The learning rate is set to 0.0001 without warm-up. The fine-tuning is conducted on Tesla V100 GPUs for approximately 10 hours on the largest corpus. Other hyper-parameters follow the settings in~\citet{lee2022formnet}.

\begin{figure}[ht!]
    \centering
    \includegraphics[width=0.85\linewidth]{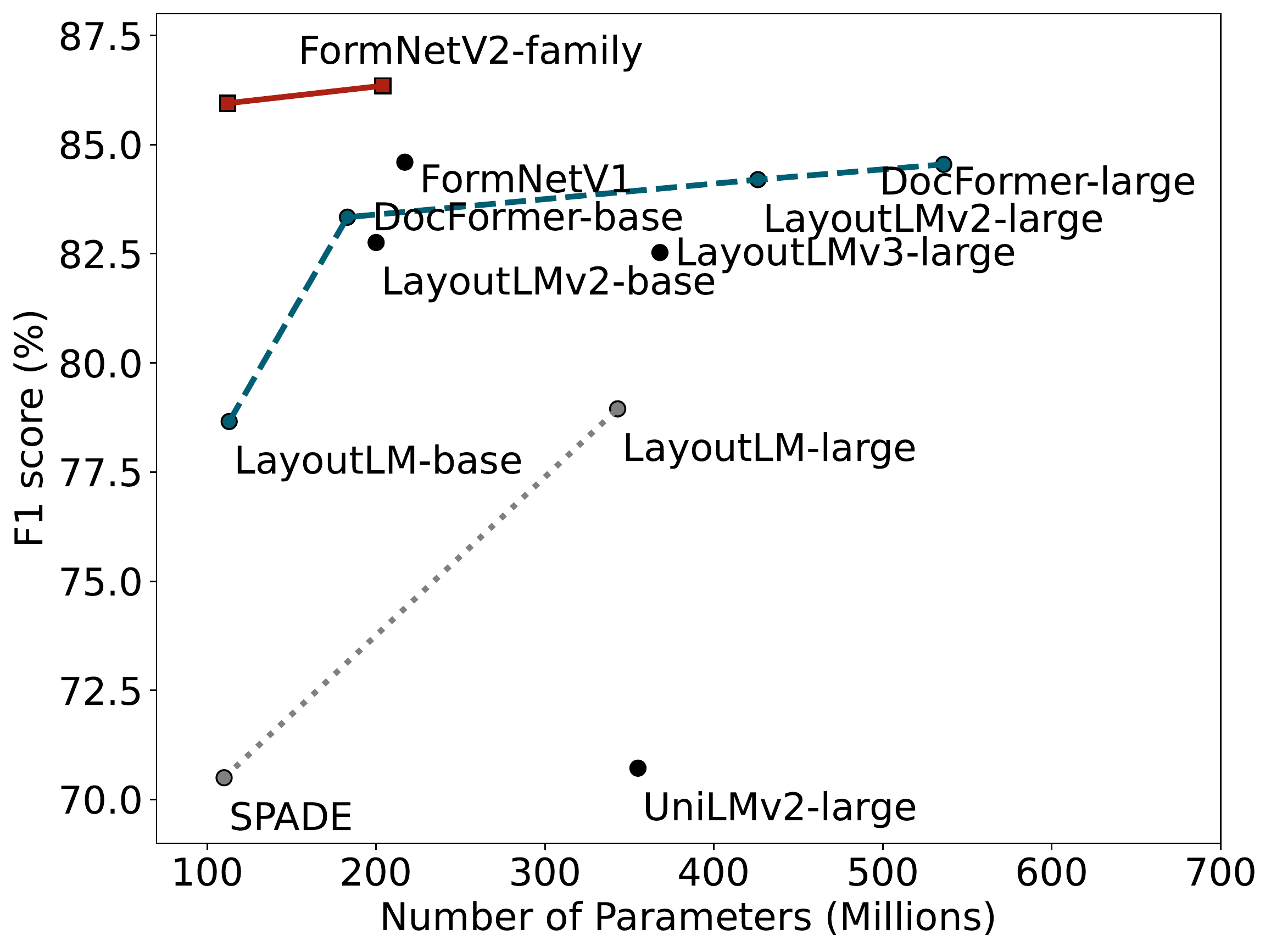}
    \vspace{-3mm}
    \caption{\textbf{Model Size vs.\ Entity Extraction F1 Score} on FUNSD benchmark. The FormNetV2 family significantly outperforms other recent approaches -- FormNetV2 achieves highest F1 score (86.35\%) while using a 2.6x smaller model than DocFormer~\citep[84.55\%;][]{appalaraju2021docformer}. FormNetV2 also outperforms FormNetV1~\cite{lee2022formnet} by a large margin (1.66 F1) while using fewer parameters.}
    \label{fig:funsd_params}
    \vspace{-5mm}
\end{figure}

\subsection{Benchmark Results}
\label{sec:benchmark}
Table~\ref{overall_comparison} lists the results that are based on the same evaluation protocal\footnote{Micro-F1 for FUNSD, CORD, and SROIE by following the implementation in~\citet{xu2020layoutlmv2}; macro-F1 for Payment~\cite{majumder2020representation}.}.

As the field is actively growing, researchers have started to explore incorporating additional information into the system. For example, LayoutLMv3~\cite{huang2022layoutlmv3} and StructuralLM~\cite{li2021structurallm} use segment-level layout positions derived from ground truth entity bounding boxes -- the $\{$Begin, Inside, Outside, End, Single$\}$ schema information~\cite{ratinov2009design} that determine the spans of entities are given to the model, which is less practical for real-world applications. We nevertheless report our results under the same protocol in column \gray{\textbf{F1}\textsuperscript{\textdagger}} in Table~\ref{overall_comparison}.
We also report LayoutLMv3 results without ground-truth entity segments for comparisons.

Furthermore, UDoc~\cite{gu2022unified} uses additional paragraph-level supervision returned by a third-party OCR engine EasyOCR\footnote{\url{https://github.com/JaidedAI/EasyOCR}}. Additional PubLayNet~\cite{zhong2019publaynet} dataset is used to pre-train the vision backbone. UDoc also uses different training/test splits (626/247) on CORD instead of the official one (800/100) adopted by other works. ERNIE-mmLayout~\cite{wang2022ernie} utilizes a third-party library spaCy\footnote{\url{spacy.io}} to provide external knowledge for the Common Sense Enhancement module in the system. The F1 scores on FUNSD and CORD are 85.74\% and 96.31\% without the external knowledge. We hope the above discussion can help clarify the standard evaluation protocol and decouple the performance improvement from modeling design vs.\ additional information.

\begin{figure*}
\centering
\begin{subfigure}{.2\textwidth}
  \centering
  \includegraphics[width=1\linewidth]{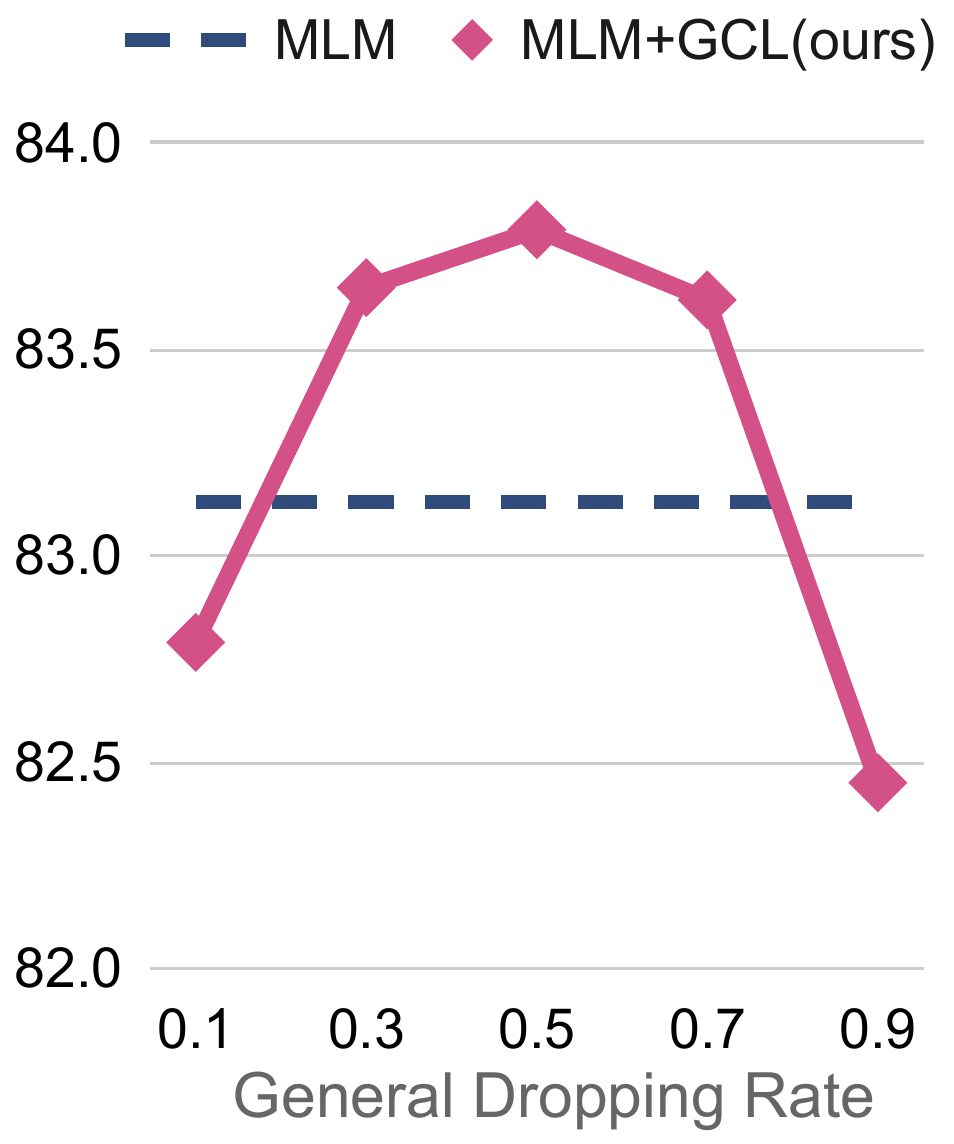}
  \caption{FUNSD}
  \label{fig:sub1}
\end{subfigure}%
\hspace{1mm}
\begin{subfigure}{.2\textwidth}
  \centering
  \includegraphics[width=1\linewidth]{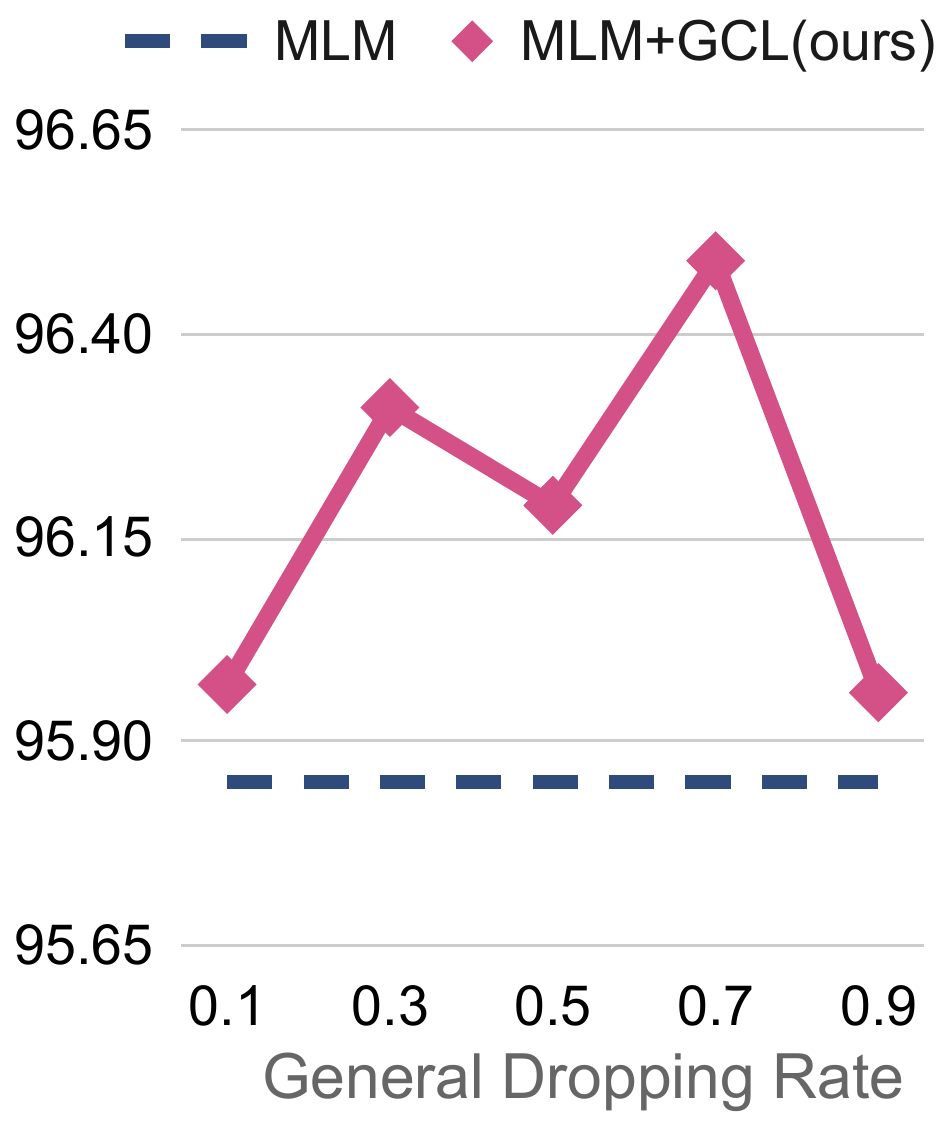}
  \caption{CORD}
  \label{fig:sub2}
\end{subfigure}
\hspace{1mm}
\begin{subfigure}{.27\textwidth}
  \centering
  \includegraphics[width=1\linewidth]{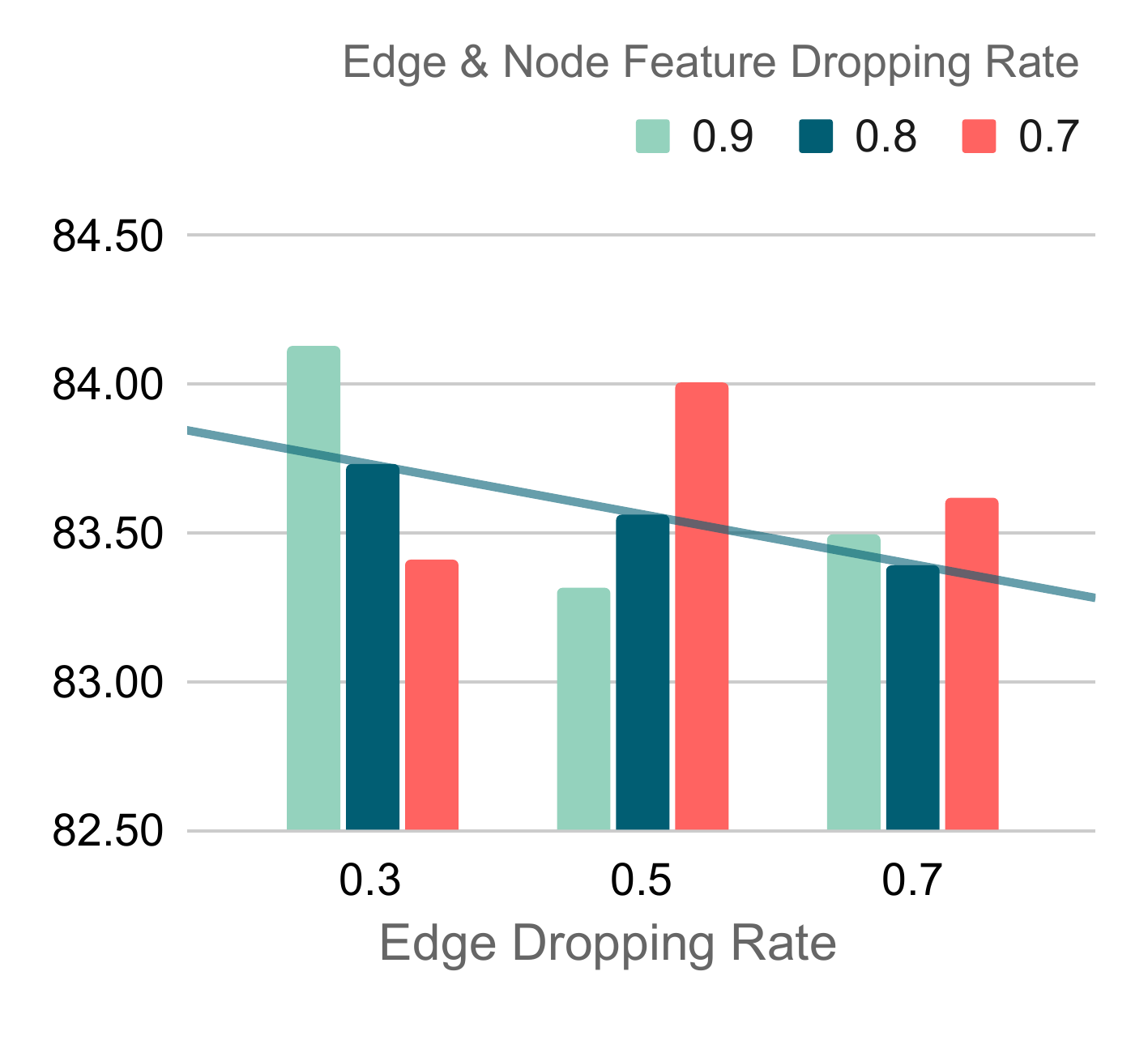}
  \caption{FUNSD}
  \label{fig:sub3}
\end{subfigure}
\hspace{1mm}
\begin{subfigure}{.27\textwidth}
  \centering
  \includegraphics[width=1\linewidth]{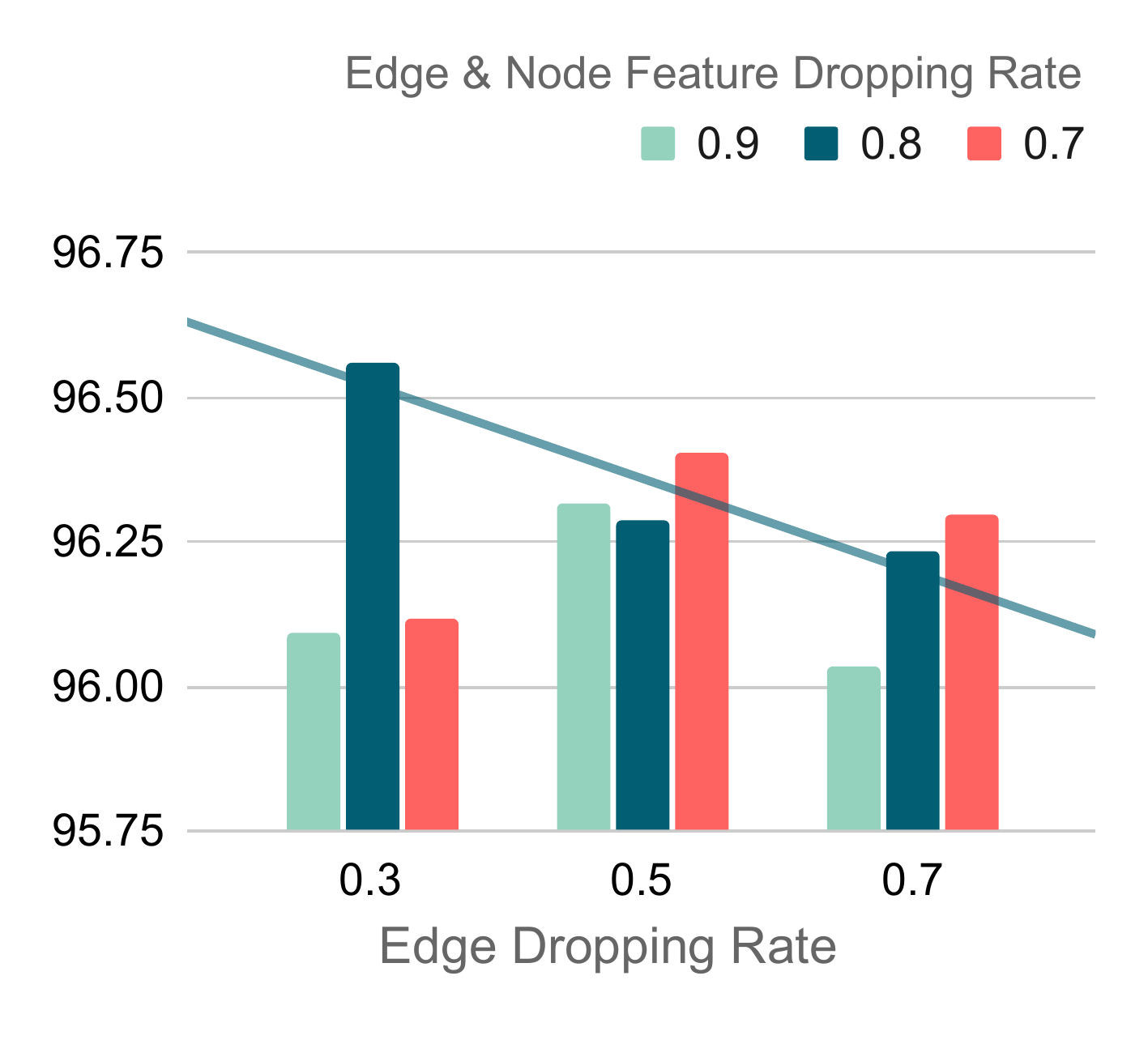}
  \caption{CORD}
  \label{fig:sub4}
\end{subfigure}
\vspace{-2mm}
\caption{\textbf{Entity Extraction F1 Score vs. Graph Corruption Mechanism} on FUNSD and CORD benchmarks. (a)(b) show results using the same drop-rate across modalities. The proposed multimodal graph contrastive learning improves MLM pretraining at almost all drop-rates; (c)(d) show results using different drop-rates across modalities. The decoupled dropping mechanism permits further boosts to the F1 scores over non-decoupled counterparts. See Sec~\ref{sec:ablation} for discussion.}
\vspace{-2mm}
\label{fig:corruption_ablation}
\end{figure*}

Figure~\ref{fig:funsd_params} shows model size vs.\ F1 score for the recent approaches that are directly comparable.
The proposed method significantly outperforms other approaches in both F1 score and parameter efficiency: FormNetV2 achieves highest F1 score (86.35\%) while using a 38\% sized model than DocFormer~\citep[84.55\%;][]{appalaraju2021docformer}. FormNetV2 also outperforms FormNetV1~\cite{lee2022formnet} by a large margin (1.66 F1) while using fewer parameters. Table~\ref{overall_comparison} shows that FormNetV2 outperforms LayoutLMv3~\cite{huang2022layoutlmv3} and StructuralLM~\cite{li2021structurallm} with a considerable performance leap while using a 55\% and 57\% sized model, respectively. From Table~\ref{overall_comparison} we also observe that using all three modalities (text+layout+image) generally outperforms using two modalities (text+layout), and using two modalities (text+layout) outperforms using one modality (text) only across different approaches.

\subsection{Ablation Studies}
\label{sec:ablation}
We perform studies over the effect of image modality, graph contrastive learning, and decoupled graph corruption. The backbone for these studies is a 4-layer 1-attention-head GCN encoder followed by a 4-layer 8-attention-head ETC transformers decoder with 512 hidden units. The model is pre-trained on the 1M IIT-CDIP subset. All other hyperparameters follow Sec~\ref{sec:setup}.

\vspace{-2mm}
\paragraph{Effect of Image Modality and Image Embedder.}
Table~\ref{image_embedder} lists results of FormNetV1 (a) backbone only, (b) with additional tokens constructed from image patches\footnote{We experiment with 32x32 image patch size, resulting in additional 256 image tokens to the model.}, and (c) with the proposed image feature extracted from edges of a graph. The networks are pre-trained with MLM only to showcase the impact of input with image modality.

We observe that while (b) provides slight F1 score improvement, it requires 32\% additional parameters over baseline (a). The proposed (c) approach achieves a significant F1 boost with less than 1\% additional parameters over baseline (a). Secondly, we find the performance of more advanced image embedders~\cite{he2016deep} is inferior to the 3-layer ConvNet used here, which suggests that these methods may be ineffective in utilizing image modality. Nevertheless, the results demonstrate the importance of image modality as part of the multimodal input. Next we will validate the importance of an effective multimodal pre-training mechanism through graph contrastive learning.


\vspace{-2mm}
\begin{table}[!ht]
\setlength{\tabcolsep}{6pt} 
\centering
\resizebox{0.49\textwidth}{!}{
\footnotesize{
\begin{tabular}{lccr}
\toprule
\textbf{Method} & \textbf{FUNSD} & \textbf{CORD} & \textbf{\#Params}  \\
\toprule
FormNetV1 & 82.53 & 95.16 & 81.7M  \\
FormNetV1+Image Patch& 82.65 & 95.43  & 107.0M  \\
FormNetV1+Edge Image (ours) & 83.13 & 95.85 & 82.3M  \\
\bottomrule
\end{tabular}
}
}
\vspace{-3mm}
\caption{\label{image_embedder} F1 with different image modality setups.} 
\vspace{-4mm}
\end{table}

\paragraph{Effect of Graph Contrastive Learning.}
The graph corruption step (Figure~\ref{fig:multimodal_gcl}) in the proposed multimodal graph contrastive learning requires corruption of the original graph at both topology and feature levels. Considering the corruption happens in multiple places: edges, edge features, and node features, a naive graph corruption implementation would be to use the same drop-rate value everywhere. In Figure~\ref{fig:corruption_ablation}(a)(b), we show the downstream entity extraction F1 scores on FUNSD and CORD datasets by varying the dropping rate value during the graph contrastive pre-training. The selected dropping rate is shared across all aforementioned places.

Results show that the proposed multimodal graph contrastive learning works out of the box across a wide range of dropping rates. It demonstrates the necessity of multimodal corruption at both topology level and feature level -- it brings up to 0.66\% and 0.64\% F1 boost on FUNSD and CORD respectively, when the model is pre-trained on MLM plus the proposed graph contrastive learning over MLM only. Our method is also stable to perturbation of different drop-rates. 

We observe less or no performance improvement when extreme drop-rates are used; for example, dropping 10\% edges and features or dropping 90\% edges and features. Intuitively, dropping too few or too much information provides either no node context changes or too few remaining node contexts in different corrupted graphs for effective contrastive learning.



\begin{figure*}
\centering
\begin{subfigure}{.45\textwidth}
  \centering
  \includegraphics[width=1\linewidth]{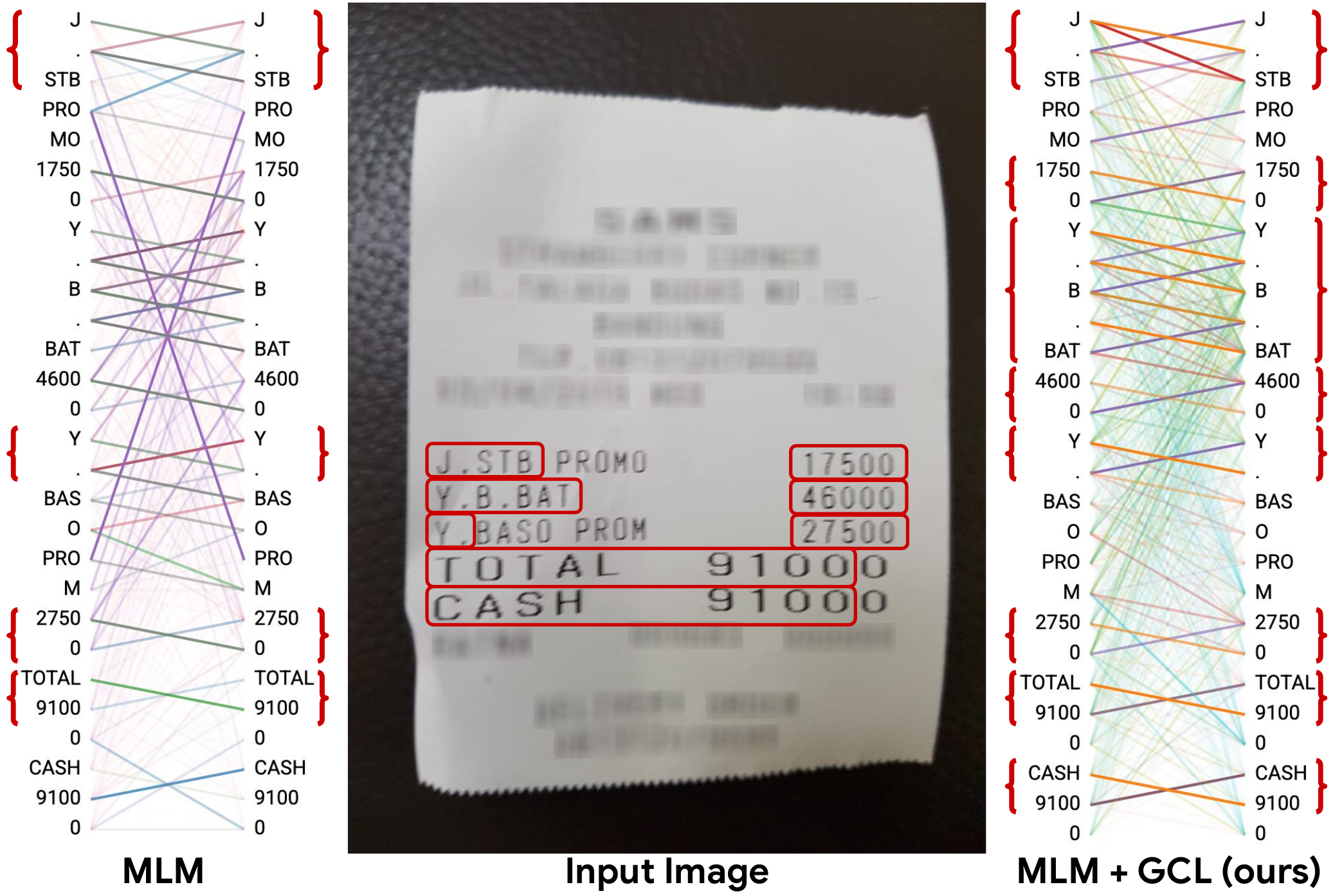}
  \caption{Attention scores w/ and w/o GCL}
  \label{fig:sub5}
\end{subfigure}%
\hspace{3mm}
\begin{subfigure}{.52\textwidth}
  \centering
  \includegraphics[width=1\linewidth]{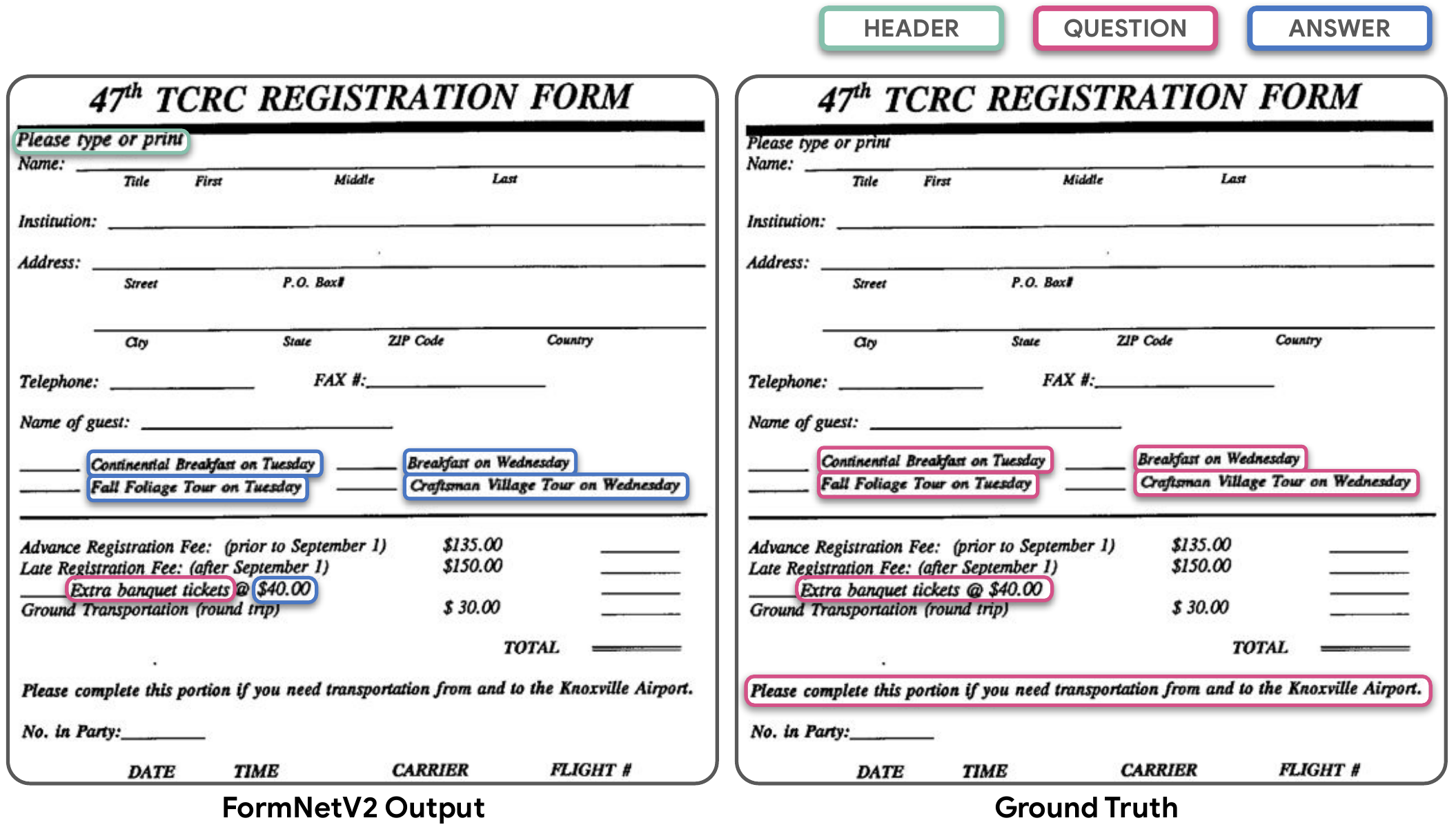}
  \caption{Model outputs for difficult cases.}
  \label{fig:sub6}
\end{subfigure}
\vspace{-2mm}
\caption{(a) The attention scores for MLM and MLM+GCL(Graph Contrastive Learning) models on CORD before fine-tuning. When pre-trained with the proposed GCL, the model can identify more meaningful token clusterings, leveraging multimodal input effectively; (b) Difficult cases where the model predictions do not match the human-annotated ground truth. In this
visualization we highlight disagreements only.}
\vspace{-1mm}
\label{fig:att_output}
\end{figure*}

\vspace{-2mm}
\paragraph{Effect of Decoupled Graph Corruption.}
In this study, we investigate whether decoupling the drop-rate in different places of graph corruption can learn better representations during pre-training and bring further improvement to the downstream entity extraction tasks.
Specifically, we select different dropping rates for all four different places: edge, layout and image features at edge level, and text features at node level. At feature level (layout, image, text), when one of the corrupted graphs selects dropping rate $p$ for a certain feature, the other corrupted graph will use the complement of the selected dropping rate $1-p$ for the same feature as introduced in Sec~\ref{alg:gcl}. 
This inductive multimodal contrastive design creates stochastically imbalanced information access to the features between two corrupted views. It provides more diverse contexts at node level in different views and makes the optimization of the contrastive objective harder, ideally generating more semantically meaningful representations between the three modalities.

Figure~\ref{fig:corruption_ablation}(c)(d) show the downstream entity extraction F1 scores on FUNSD and CORD datasets by pre-training with three different edge dropping rates and three different feature dropping rates.
We observe that decoupling the dropping rate at various levels further boosts the performance on both datasets -- it brings another 0.34\% and 0.07\% F1 boost on FUNSD and CORD respectively, when decoupled dropping rates are used over the non-decoupled ones.

We also observe nonlinear interactions between different dropping rates at edge level and feature level. The best performing feature dropping rate might be sub-optimal when a different edge dropping rate is applied. This is noteworthy but not surprising behavior, since different edge dropping rates would drastically change the graph topology (and therefore the node embeddings). We expect the amount of information needed for maximizing the agreement of node contexts between two corrupted graphs to be different when the graph topology is altered. Nevertheless, we find that low edge dropping rates (e.g. 0.3) generally perform better than high edge dropping rates, and therefore select a low edge dropping rate in our final design.



\vspace{-2mm}
\paragraph{Visualization.}
We visualize~\citep{vig-2019-multiscale} the local-to-local attention scores of a CORD example for model pre-trained with MLM only and MLM+GCL but before fine-tuning in Figure~\ref{fig:att_output}(a). We observe that with GCL, the model can identify more meaningful token clusterings, leveraging multimodal input more effectively. 

We also show sample model outputs that do not match the human-annotated ground truth in Figure~\ref{fig:att_output}(b). The model confuses between `header` and `other` on the top of the form and between `question` and `answer` for the multiple choice questions on the bottom half of the form. More visualization can be found in Figure~\ref{fig:output_visual_appendix} in Appendix.

\section{Conclusion}
FormNetV2 augments an existing strong FormNetV1 backbone with image features bounded by pairs of neighboring tokens and the graph contrastive objective that learns to differentiate between the multimodal token representations of two corrupted versions of an input graph. The centralized design sheds new light to the understanding of multimodal form understanding.

\section{Limitations}
Our work follows the general assumption that the training and test set contain the same list of predefined entities. Without additional or necessary modifications, the few-shot or zero-shot capability of the model is expected to be limited. Future work includes exploring prompt-based architectures to unify pre-training and fine-tuning into the same query-based procedure. 

\section{Ethics Consideration}
We have read and compiled with the ACL Code of Ethics. 
The proposed FormNetV2 follows the prevailing large-scale pre-training then fine-tuning framework. 
Although we use the standard IIT-CDIP dataset for pre-training in all experiments, the proposed method is not limited to using specific datasets for pre-training. 
Therefore, it shares the same potential concerns of existing large language models, such as biases from the pre-training data and privacy considerations. We suggest following a rigorous and careful protocol when preparing the pre-training data for public-facing applications.


\bibliography{anthology,custom}

\clearpage

\appendix

\section{Appendix}
\label{sec:appendix}

\subsection{Image Embedder Architecture}
\label{app:image_embedder}
Our image embedder is a 3-layer ConvNet with filter sizes \{32, 64, 128\} and kernel size 3 throughout. Stride 2 is used in the middle layer and stride 1 is used everywhere else. We resize the input document image to 512$\times$512 with  aspect ratio fixed and zero padding for the background region. After extracting the dense features of the whole input image, we perform feature RoI pooling~\cite{he2017mask} within the bounding box that joins a pair of tokens connected by a GCN edge. The height and width of the pooled region are set to 3 and 16, respectively. Finally, the pooled features go through another 3-layer ConvNet with filter size \{64, 32, 16\} and kernel size 3 throughout. Stride 2 is used in the first 2 layers horizontally and stride 1 is used everywhere else. To consume image modality in our backbone model, we simply concatenate the pooled image features with the existing layout features at edge level of GCN as shown in Figure~\ref{fig:multimodal_feat}.

\subsection{More Implementation Details}
\label{app:v3_ablation}
We conduct additional experiments\footnote{github.com/Jyouhou/unilm-test} on FUNSD and CORD using base and large versions of LayoutLMv3~\cite{huang2022layoutlmv3}. Instead of using entity segment indexes inferred from ground truth, we use word boxes provided by OCR. We observe considerable performance degradation when the model has access to word-level box information instead of segment-level. The results are shown in Table~\ref{layoutlmv3}.

\begin{table}[!ht]
\setlength{\tabcolsep}{6pt} 
\centering
\resizebox{0.48\textwidth}{!}{
\footnotesize{
\begin{tabular}{llcc}
\toprule
\textbf{Method} & \textbf{Setting} & \textbf{FUNSD} & \textbf{CORD}  \\
\toprule
LayoutLMv3-base & Reported   & 90.29 & 96.56  \\
                & Reproduced & 90.59 & 95.85  \\
                & Word box   & 78.35 & 95.81  \\
\cmidrule{1-4}                
LayoutLMv3-large & Reported   & 92.08 & 97.46  \\
                 & Reproduced & 92.14 & 96.78  \\
                 & Word box   & 82.53 & 95.92  \\                
\bottomrule
\end{tabular}
}
}
\vspace{-3mm}
\caption{\label{layoutlmv3} LayoutLMv3 results with entity segment indexes (reproduced) or word level indexes (word box). We observe considerable performance degradation when the model has access to word-level box information instead of segment-level.} 
\vspace{-4mm}
\end{table}

\subsection{Preliminaries}
\label{app:preliminaries}

FormNetV1~\cite{lee2022formnet} simplifies the task of document entity extraction by framing it as fundamentally text-centric, and then seeks to solve the problems that immediately arise from this. Serialized forms can be very long, so FormNetV1 uses a transformer architecture with a local attention window (ETC) as the backbone to work around the quadratic memory cost of attention. This component of the system effectively captures the text modality.

OCR serialization also distorts strong cues of semantic relatedness -- a word that is just above another word may be related to it, but if there are many tokens to the right of the upper word or to the left of the lower word, they will intervene between the two after serialization, and the model will be unable to take advantage of the heuristic that nearby tokens tend to be related. To address this, FormNetV1 adapts the attention mechanism to model spatial relationships between tokens using Rich Attention, a mathematically sound way of conditioning attention on low-level spatial features without resorting to quantizing the document into regions associated with distinct embeddings in a lookup table. This allows the system to build powerful representations from the layout modality for tokens that fall within the local attention window.

Finally, while Rich Attention maximizes the potential of local attention, there remains the problem of what to do when there are so many interveners between two related tokens that they do not fall within the local attention window and cannot attend to each other at all. To this end FormNetV1 includes a graph convolutional network (GCN) contextualization step \emph{before} serializing the text to send to the transformer component. The graph for the GCN locates up to $K$ potentially related neighbors for each token before convolving to build up the token representations that will be fed to the transformer after OCR serialization. Unlike with Rich Attention, which directly learns concepts like ``above'', ``below'', and infinitely many degrees of ``nearness'', the graph at this stage does not consider spatial relationships beyond ``is a neighbor'' and ``is not a neighbor'' -- see Figure~\ref{fig:graph_construction}. This allows the network to build a weaker but more complete picture of the layout modality than Rich Attention, which is constrained by local attention. A similar architecture is also found to be useful in graph learning tasks by~\citet{wu2021representing}.

\begin{figure*}[ht!]
    \centering
    \includegraphics[width=0.98\linewidth]{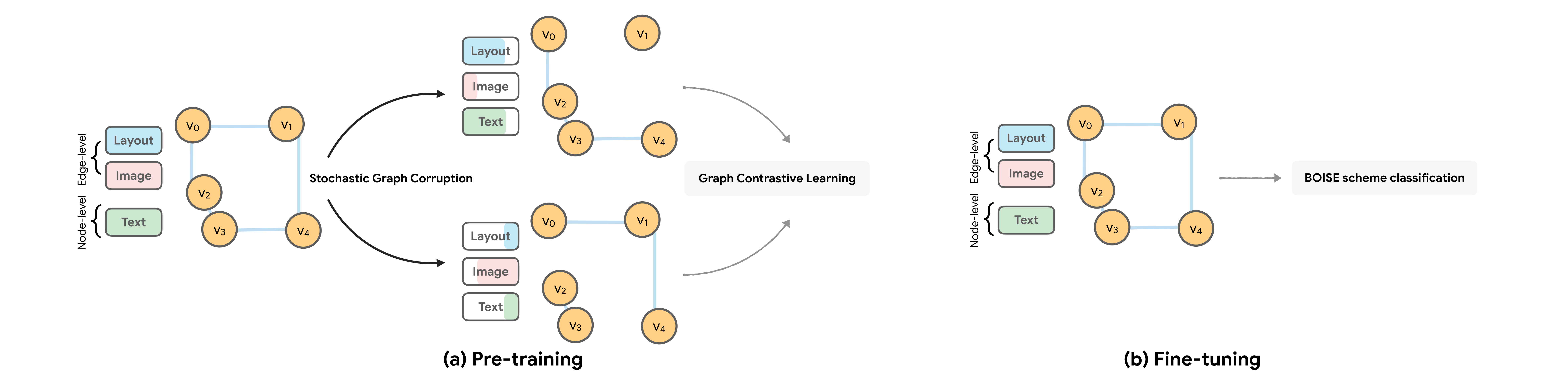}
    \vspace{-2mm}
    \caption{(a) During multimodal graph contrastive pre-training, two corrupted graphs are sampled from an input graph by corruption of graph topology (edges) and attributes (multimodal features). (b) During task-specific fine-tuning, only the original input graph is used.}
    \label{fig:training_phases}
    \vspace{2mm}
\end{figure*}

\begin{figure*}[th!]
    \centering
    \includegraphics[width=0.95\linewidth]{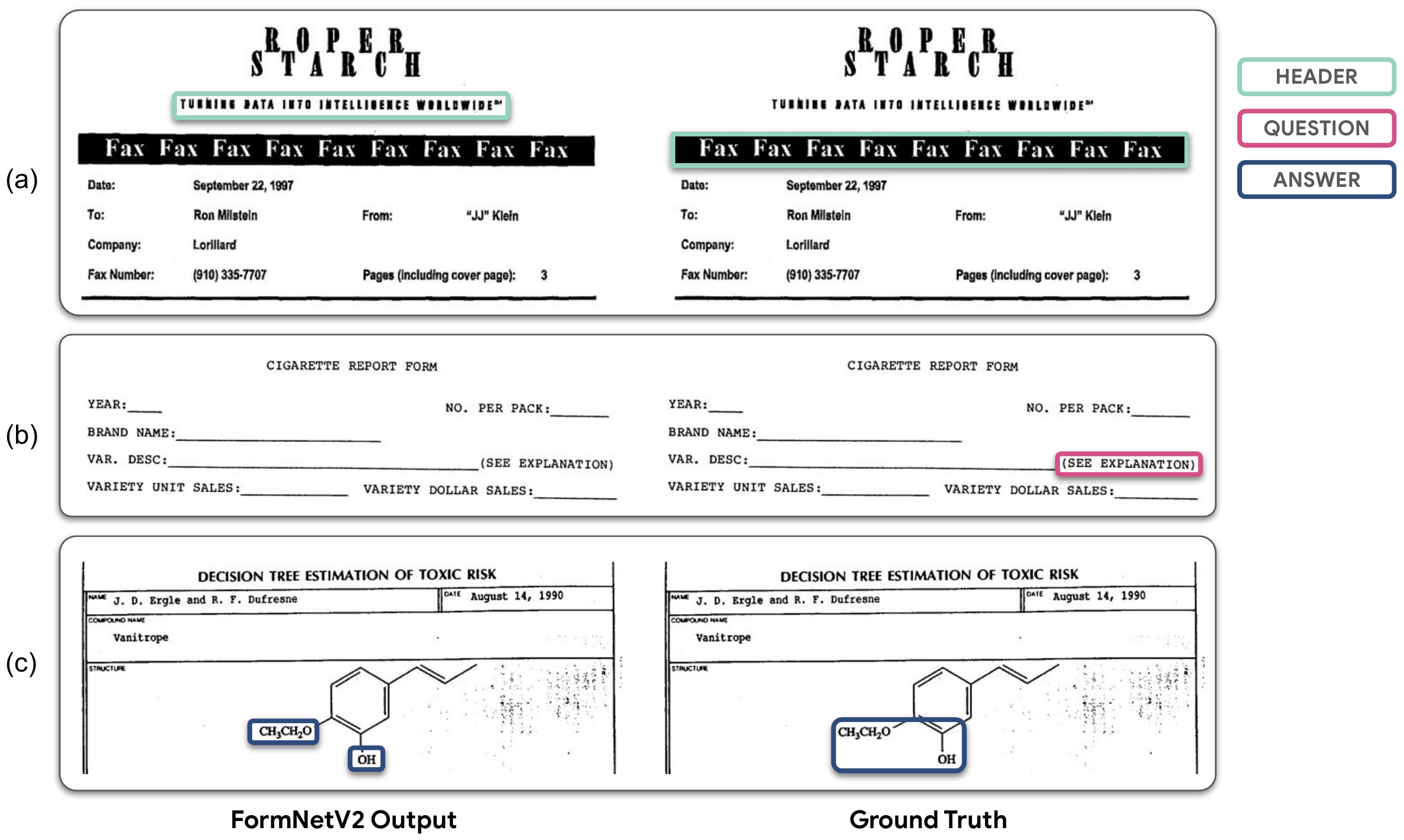}
    \vspace{-2mm}
    \caption{The ambiguous cases where the model predictions do not match the human-annotated ground truth. In this visualization we only showcase mismatched entities.}
    \label{fig:output_visual_appendix}
\end{figure*}

Thus the three main components of FormNetV1 cover each other's weaknesses, strategically trading off representational power and computational efficiency in order to allow the system to construct useful representations while simplifying the problem to be fundamentally textual rather than visual.
The final system was pretrained end-to-end on large scale unlabeled form documents with a standard masked language modeling (MLM) objective.

\subsection{Output Visualization}
\label{app:output_visualization}
Figure~\ref{fig:output_visual_appendix} shows additional FormNetV2 model outputs on FUNSD.

\subsection{License or Terms}
Please see the license or terms for IIT-CDIP\footnote{ir.nist.gov/cdip/README.txt}, FUNSD\footnote{guillaumejaume.github.io/FUNSD/work/}, CORD\footnote{github.com/clovaai/cord/blob/master/LICENSE-CC-BY}, and SROIE\footnote{rrc.cvc.uab.es/?ch=13} in the corresponding footnotes.

\end{document}